\documentclass[10pt,twocolumn,letterpaper]{article}

\usepackage{caption} 
\usepackage{authblk}
\usepackage{wacv}
\usepackage{times}
\usepackage{epsfig}
\usepackage{graphicx}
\usepackage{amsmath}
\usepackage{amssymb}
\usepackage{booktabs}
\usepackage{enumitem}
\usepackage{adjustbox}
\usepackage{placeins}
\usepackage{subfiles}
\usepackage{txfonts}

\usepackage{spconf}

\usepackage[utf8]{inputenc}

\wacvalgorithmstrack   
\wacvfinalcopy 

\usepackage[pagebackref=true,breaklinks=true,colorlinks,bookmarks=false]{hyperref}

\begin{document}

\title{Guided Attention for Next Active Object \\ @ Ego4D Short Term Object Interaction Anticipation Challenge }


\name{Sanket Thakur\textsuperscript{1,4}, Cigdem Beyan\textsuperscript{2,1}, Pietro Morerio\textsuperscript{1}, Vittorio Murino\textsuperscript{3,1}, Alessio Del Bue\textsuperscript{1}}
\address{\textsuperscript{1} Pattern Analysis and Computer Vision, Istituto Italiano di Tecnologia, Genoa, Italy \\
\textsuperscript{2} University of Trento, Trento, Italy \\
\textsuperscript{3} University of Verona, Verona, Italy \\
\textsuperscript{4} DITEN, University of Genoa, Genoa, Italy \\  \hline
\\
\textbf{\{sanket.thakur, pietro.morerio, vittorio.murino, alessio.delbue\}@iit.it, cigdem.beyan@unitn.it}} 

\maketitle
\thispagestyle{empty}


\begin{abstract}
In this technical report, we describe the Guided-Attention mechanism \cite{gano} based solution for the short-term anticipation (STA) challenge for the EGO4D challenge. It combines the object detections, and the spatiotemporal features extracted from video clips, enhancing the motion and contextual information, and further decoding the object-centric and motion-centric information to address the problem of STA in egocentric videos. For the challenge, we build our model on top of StillFast \cite{ragusa2023stillfast} with Guided Attention applied on fast network. Our model obtains better performance on the validation set and also achieves state-of-the-art (SOTA) results on the challenge test set for EGO4D Short-Term Object Interaction Anticipation Challenge.
\end{abstract}
%

\section{Introduction}
\label{sec:intro}

Short-term action anticipation in egocentric videos is the task of predicting the actions that are likely to be performed by a first-person in the near future, along with foreseeing a next-active-object interaction and an estimate of the time at which the interaction will occur. The computer vision community has gathered significant progress in the field of action anticipation in egocentric videos, which predicts only the action labels \cite{avt,liu2019forecasting,rulstm, memvit2022}. However, the use of the next active objects \cite{anacto,furnari2017next,ADL} has not been widely explored in the current literature. Recently \cite{ego4d} proposed the use of next active objects for anticipating future actions. Based on the description of \cite{ego4d}, the task of short-term anticipation remains challenging since it requires the ability to anticipate both the mode of action and the time at which the action will begin, known as the time to contact.

The next active objects play a crucial role in understanding the nature of interactions happening in a video. They provide important context for predicting future actions as they indicate which objects are likely to be involved in the next action \cite{tpami_contact}. In this vein, we propose a novel approach for addressing the problem of STA in egocentric videos. Our approach utilizes a guided attention mechanism between the spatiotemporal features extracted from video clips and objects to enhance the spatial object-centric information as proposed in \cite{gano}. Our model builds on top of StillFast \cite{ragusa2023stillfast}. 

The main contribution of this paper is to show the importance of the proposed guided attention mechanism for the next active object-based STA. Our approach aims to better capture the visual cues related to the next active objects, which we assume are highly correlated with the action that will follow.
The proposed GANO model is trained and evaluated on the largest egocentric video dataset: Ego4D \cite{ego4d}. Experimental results demonstrate that $GANO_{v2}$ outperforms the state-of-the-art (SOTA) egocentric action anticipation methods. Additionally, we refer the reader to \cite{gano} which investigates the impact of guided attention on the performance of the GANO model for transformer-based prediction heads on ``v1'' of the EGO4D dataset. The results justify that incorporating guided attention, in other words, combining the information from spatiotemporal features and objects, improves the STA performance.

\section{Our Approach}
\label{sec:pagestyle}
We now describe the details of our method, $GANO_{v2}$. However, we refer the readers to the original paper \cite{gano} for more details on Guided-Attention. 

\begin{figure*}[t!]
\centering
\includegraphics[width=\linewidth, height=0.55\linewidth]{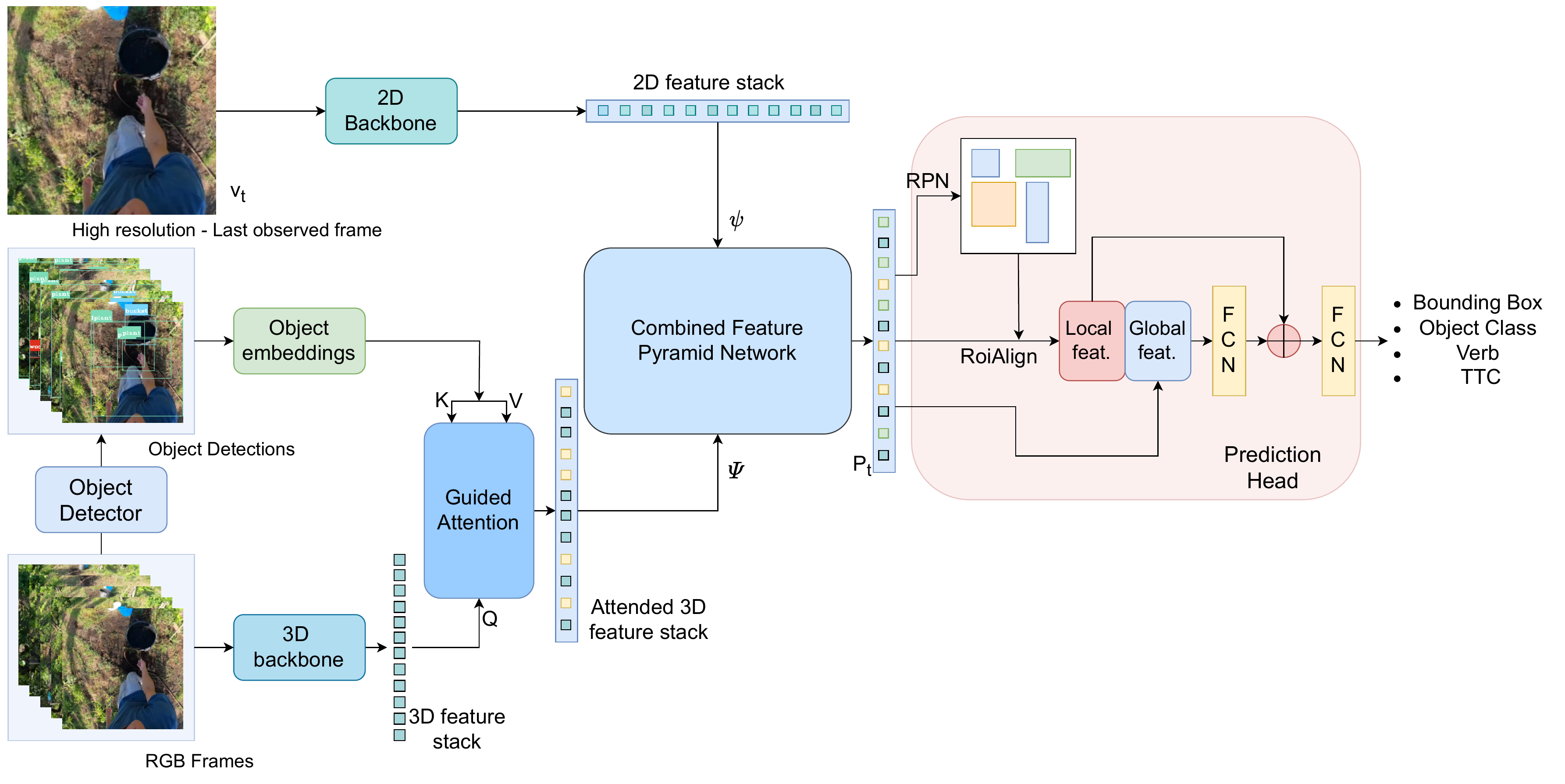}
\caption{Our $GANO_{v2}$ model uses a low-resolution video clip with sampled frames and a high-resolution target frame. Object detections are extracted for sampled input frames and are fused with patch features using a multi-head attention layer. The resulting attended 3D feature stack is merged with the 2D feature stack using a feature pyramid network and followed by a prediction head. The prediction head uses an RPN network to generate local feature which is fused with global features from $P_t$, with a Global Average Pooling operation, and concatenated with local features. These features are fed into a fusion network and then summed to the original local features through residual connections. The local-global representations are then used to predict the final prediction for NAO bounding boxes, object class, verb, and TTC. 
}
\label{fig:method}
\end{figure*}

\subsection{Backbone}
Given an input video clip, the proposed model takes as input a high-resolution last observed frame from the clip and low-resolution sampled video, $V = \{{v_i\}^T_{i=1}}$ where $v_i \in {R}^{C\times{H_o}\times{W_o}}$. An object detector \cite{fastercnn} pre-trained on \cite{ego4d} is used to extract object detections for each sampled video frame. The detection consists of the bounding boxes $(x1, y1, x2, y2)$ along with their class label. 
To process the input image and video simultaneously, the proposed model comprises a two-branch backbone.

A 2D CNN backbone processes the high-resolution frame $v_T$ and produces a stack of 2D features at different spatial resolutions, $\psi$. 
The ``fast'' branch consists of two parts:  (1) A 3D CNN backbone processes the video, V, and outputs a stack of 3D features. (2) In parallel, an MLP is employed to generate object embeddings from the object detections (class label $, x1, y1, x2, y2$) for the input video frames.  
In the final process, the stack of 3D features is fused with object embeddings using the Guided-Attention approach. 

\subsection{Object Guided Attention.}
We use Objects-Guided Multi-head Attention to efficiently fuse spatiotemporal information across the video clip, and object detections and then infer long-term dependencies across both. Using a single attention head does not suffice as our goal is to allow detection embeddings to attend to co-related patches from the video clip. Therefore, we modify the Multi-Head Attention described in \cite{vaswani_attn} in a way that it can take the inputs from both modalities. To do so, we set Query $Q$, Key $K$, and Value $V$ as follows.

\begin{align}
    \begin{array}{c}
     Q = f_{vid}(F_i), \text{where}\;i \in [1,..N], \\
     \\
    K, V = f_{obj}(O_j), \text{where}\;j \in [1,...M],\\
    \\
    \text{Object-Guided Attention(Q,K,V)} = Concat(h_1, ... h_h)W_o, \\
    \text{where}\; h_i = Attention({QW_i}^Q, {KW_i}^K, {VW_i}^V), \\
    \\
    \text{and Attention(Q, K, V)} = softmax(\dfrac{{{QK}^T}}{d_k})V  
    \\
    \end{array}  
\end{align}

where ${W_i}^Q, {W_i}^K, and {W_i}^V$ are learnable parameter matrices and $d_k$ represents the dimensions of $K$. The output of this Object-Guided Multi-Attention is the attended features for the provided object embeddings, denoted as $F_i$ for a single feature layer $i$. The entire 3D feature stack, $\Psi$, ($\Psi = \{{F_i\}^N_{i=1}}$) for N feature layers, is sent to the Combined Feature Pyramid Network. \\

\begin{table*}[t]
\resizebox{\linewidth}{!}{
\begin{tabular}{|l|ccccc|} \hline
Models & Data Split & Noun & N+V & N+TTC & Overall  \\  \hline
FRCNN+SF. \cite{feichtenhofer2019slowfast} & val & 21.0 & 7.45 & 7.04 & 2.98  \\
StillFast \cite{ragusa2023stillfast} & val & 20.26 & 10.37 & 7.16 & 3.96  \\
$GANO_{v2}$ (Ours) & val & \textbf{20.52} & \textbf{10.42} & \textbf{7.28} & \textbf{3.99}  \\ \hline
FRCNN+SF. \cite{feichtenhofer2019slowfast} & test & 26.15 & 9.45 & 8.69 & 3.61  \\
StillFast \cite{ragusa2023stillfast} & test & 25.06 & 13.29 & \textbf{9.14} & 5.12  \\
$GANO_{v2}$ (Ours) & test & \textbf{25.67} & \textbf{13.60} & 9.02 & \textbf{5.16}  \\ \hline
\end{tabular}}
\caption{Results$\%$ in Top-5 mean Average Precision on the validation and test sets of EGO4D v2. In the header of the table, N+V stands for Noun + Verb and N+TTC stands for Noun + Time to Contact. Best results per column within a section of comparable results (horizontal lines) are reported in bold}
\label{table:results}
\vspace{-7pt}
\end{table*}

\begin{table}
\resizebox{\linewidth}{!}{
\begin{tabular}{|l|cccc|} \hline
Guided Fusion in Layer & Noun & N$+$V & N$+$TTC & Overall  \\  \hline
1 & 18.7 & 9.42 & 6.27 & 3.22 \\
4 & 20.47 & 10.40 & 7.20 & 3.96 \\
All & \textbf{20.52} & \textbf{10.42} & \textbf{7.28} & \textbf{3.99} \\ \hline
\end{tabular}}
\caption{Guided Attention fusion prediction for each output layer of 3D CNN. Results$\%$ in Top-5 mean Average Precision on the validation set of EGO4D v2.}
\label{table:fusion_layer}
\end{table}

\subsection{Feature Pyramid Network and Prediction Head} 
We adopt the Combined Feature Pyramid Layer and Predicting head from \cite{ragusa2023stillfast} for the purpose of fusing 2D and 3D feature stacks for mid-level feature fusion and final prediction respectively. The 3D feature maps, $\Psi$ are interpolated and averaged out temporally to match the shape of $\psi$, followed by a 3$\times$3 convolutional layer. The resulting features summed to the 2D features, $\psi$, and then passed through another 3x3 convolutional layer. The resulting feature maps are then fed to a standard Feature Pyramid Layer \cite{FPN}.  

The prediction head is based on Detectron2 \cite{wu2019detectron2} implementation. It consists of a Region Proposal Network (RPN) which predicts region proposals from the feature pyramid. A RoiAlign layer is then used to extract local features from the region proposals. As mentioned in \cite{ragusa2023stillfast}, a global average pooling from the feature pyramid is also applied to the final layer of feature pyramid outputs and concatenated with local features from region proposals, to be followed by a dense layer. The resulting representations are summed to the original local features through a residual connection. The final features are then used to predict the object class, bounding boxes, verbs, and TTC.  We refer readers to \cite{ragusa2023stillfast} for further details. 

\subsection{Training and Implementation details} 
The model is trained end-to-end using classification and regression loss for verb and TTC prediction. In addition, we also employ the standard faster-RCNN losses. We performed experiments on the large-scale egocentric dataset EGO4D \cite{ego4d}. 
We preprocess the input video clips by randomly scaling the height between 248 and 280px and taking 224px crops at training time. We sample 32 consecutive frames as input for the low-resolution stream. The object detections are extracted on the original ``high'' resolution frame and are then scaled down to match the input shape of video frames. In our experiment, we use a ResNet-50 as 2D CNN and an X3D-M as 3D CNN. $GANO_{v2}$ was trained with an SGD optimizer for 20 epochs with a cosine learning rate of $1e-5$ with a batch size of 4 and a weight decay of $1e-6$ on two NVIDIA-SMI Tesla V100 GPU.

\noindent
\subsection{Results.}
 The results in Table \ref{table:results} demonstrate that $GANO_{v2}$ outperforms all baseline methods across all metrics evaluated on ``v2'' of the EGO4D dataset. We also conducted an ablation study in Table \ref{table:fusion_layer} to investigate the impact of the \emph{guided attention} mechanism on different feature layer(s) of the output of 3D CNN, $\Psi$. It is noted that the performance improves if Multi-head attention fusion is applied on the last layer instead only on the initial feature layer of 3D CNN. However, we achieve the best performance if attention is employed to all the feature layers. 

\section{Conclusion and Limitations}
\label{sec:conclusion}
We have presented the Guided-Attention for Next Active Object v2 ($GANO_{v2}$ architecture as used in the EGO4D 2023 challenge. We propose an end-to-end architecture for predictive video tasks for short-term anticipation which involves predicting the next-active-object class, its location (bounding box), future action, and the time to contact. Our model obtains better performance as compared to other submissions on the test set of ``v2'' of the dataset. The limitation is that it relies on the performance of object detector for guided attention. In the future, we plan to improve performance by exploring different modalities and fusion-based methods.

\newpage
\bibliographystyle{IEEEbib}
\bibliography{main}

\end{document}